\documentclass[10pt,twocolumn,letterpaper]{article}

\usepackage{cvpr}
\usepackage{times}
\usepackage{epsfig}
\usepackage{graphicx}
\usepackage{amsmath}
\usepackage{amssymb}

\usepackage{hyperref}

\cvprfinalcopy 

\def\cvprPaperID{****} 

\begin{document}

\title{Dockerface: an Easy to Install and Use Faster R-CNN Face Detector in a Docker Container}

\author{
\begin{tabular}{c@{\hskip 0.5in}c}
Nataniel Ruiz & James M. Rehg \\
{\tt\small nataniel.ruiz@gatech.edu} & {\tt\small rehg@gatech.edu}
\end{tabular} \\[10pt]
Center for Behavioral Imaging\\
College of Computing\\
Georgia Institute of Technology
}

\maketitle

\begin{abstract}
Face detection is a very important task and a necessary pre-processing step for many applications such as facial landmark detection, pose estimation, sentiment analysis and face recognition. Not only is face detection an important pre-processing step in computer vision applications but also in computational psychology, behavioral imaging and other fields where researchers might not be initiated in computer vision frameworks and state-of-the-art detection applications.A large part of existing research that includes face detection as a pre-processing step uses existing out-of-the-box detectors such as the HoG-based dlib and the OpenCV Haar face detector which are no longer state-of-the-art - they are primarily used because of their ease of use and accessibility. We introduce Dockerface, a very accurate Faster R-CNN face detector in a Docker container which requires no training and is easy to install and use.
\end{abstract}

\section{Introduction} \label{sec1}
Deep learning has recently revolutionized computer vision and one of its most important applications is the ability to consistently localize and recognize objects. In particular detecting faces in images and video is useful in countless domains.

An example application is localizing faces in order to verify the identity of a person which is used in biometrics for video surveillance. Another example consists in digital cameras detecting faces for auto-focus and smile detection for automatic snapshots.

We identified two main problems that have impeded and currently impede the widespread dissemination of state-of-the-art methods for face detection:
\begin{itemize}
\item Many commercial applications for face detection have not yet transitioned to deep learning. This means that they do not perform nearly as accurately as one would like.
\item Open source state-of-the-art deep learning face detection algorithms are not easily accessible to the wider research community. Obstacles range from not including pre-trained models to having to go through extensive installation processes and software compilation to use the applications.
\end{itemize}

In this work we present the following contributions in the hopes of solving these problems and filling this niche:
\begin{itemize}
\item We replicate work done by Jiang and Miller~\cite{jiang2017face} in which they show that by training the general purpose object detection Faster R-CNN network one can obtain state-of-the-art results on the face detection task. This is tested in section \ref{sec3} by reporting results on widely used face detection benchmark datasets.
\item We train a Faster R-CNN network for the face detection task and set it up in an easy to download and install Docker container for Linux. We include straightforward scripts to detect faces in video and images and include step by step instructions for their use. We baptize our application Dockerface and release all of the code as open source including detailed instructions at \url{https://github.com/natanielruiz/dockerface}.
\item We test Dockerface on an excerpt of real first person vision videos of neurotypical children and children with autism. In section \ref{sec4} we compare results with out-of-the-box face detectors widely used in research, namely OpenCV Haar Cascade face detection, dlib detector, the commercial OMRON OKAO detector. We show that Dockerface is much more accurate than these detectors.
\end{itemize}

In summary Dockerface is a prime candidate to replace previously widely-used methods in research due to its accuracy and ease of use. We have observed that many interesting works on facial analysis have unnecessarily used a suboptimal face detector as a preprocessing step, potentially resulting in a loss of data.
For example, for the facial landmark detection task, the bounding box of the face is required to demonstrate the method. If this bounding box is inaccurate or if the detector could not detect the face then this image might be discarded and consequently the main algorithm would not be tested on these difficult examples. In another example, the automatic analysis of children's social behaviors~\cite{Rehg2014,ye2015detecting} often begins with the detection of their faces.

\section{Faster R-CNN in a Docker Container} \label{sec2}

Here we describe the obstacles that prevent widespread use of state-of-the-art face detectors by researchers in a broad range of fields, and our solution based on Docker.

\subsection{Obstacles to Painless Deployment of Faster R-CNN in Caffe} \label{sec2-1}
Jiang and Learned-Miller demonstrated in \cite{jiang2017face} that by training the Faster R-CNN~\cite{ren2015faster} model on the task of face detection it was possible to achieve very good results on existing benchmark datasets. The Faster R-CNN model was implemented in MATLAB and ported by the authors to the Caffe framework which uses a Python interface.

There are several obstacles in using the Faster R-CNN model, most of these issues are encountered by researchers that lack advanced computer skills, and, even if they are initiated do not have the time to commit substantial effort to surmount these issues since detecting faces is just a preprocessing step required for their research and is not central to the idea that they want to develop. The obstacles are the following:
\begin{itemize}
\item The deep learning framework Caffe has a long list of software and library requirements for installation. These are automatically installed in our package.
\item Caffe has to be compiled before being used which in some cases might necessitate linking the Makefile to all of the required binaries and libraries. All of these configurations are automatically done in our package and compilation is painless.
\item Using a deep learning framework requires installation of NVIDIA drivers and the NVIDIA CUDA \cite{kirk2007nvidia} software for Graphics Processing Unit (GPU) Parallel Computing. While our software package cannot circumvent the necessity of manually installing them it guarantees that they are linked correctly to the Caffe framework and provides easy compilation with GPU support.
\item Processing video in Python is a fastidious task. The best and most popular library that supports video processing is OpenCV. Yet it has to be compiled from source specifically with video support. Compilation is long and if not carefully done can result in conflicts. OpenCV with video support is automatically compiled in our package.
\item The Faster R-CNN model has to be trained. Jiang and Learned-Miller provide convenient training scripts which allow us to retrain our own model and which is supplied in our package.
\item Scripts have to be written by the user, who might not be accustomed to Python programming, to process image and video using the Faster R-CNN. These scripts are provided in our package, easy to use and have text outputs and image or video outputs.
\end{itemize}

\subsection{Solution: Build a Docker Container} \label{sec2-2}
In order to build an accessible and easily deployable face detector, we turn to Docker~\cite{merkel2014docker} which provides lightweight linux containers for easy deployment. These containers are hardware-agnostic when using CPU applications but since we need to use NVIDIA GPUs we used a convenient Docker plugin called nvidia-docker which makes the image agnostic to the NVIDIA driver which allows the user to install their supported NVIDIA driver once on their machine and use the container without fully reinstalling the exact same driver on it~\cite{nvidia-docker}.

First we need to define a Docker container with a Dockerfile. A Dockerfile is a text file which includes all of the different steps that go into building the container which are carried out sequentially when it is time to build it. We start out with an NVIDIA parent image which contains CUDA 8.0 and cuDNN v5. We then install all requirements for the Caffe framework and download the OpenCV library. Subsequently we download our fork of the Faster R-CNN repository which includes our inference scripts. We download the face detection model weights that have been trained on the WIDER dataset. Finally we download a merged branch of the latest Caffe framework with the Faster R-CNN layers which we compile with cuDNN support.

We build the application using this Dockerfile specification and we share the image with the public at \url{https://hub.docker.com/r/natanielruiz/dockerface/} - detailed and up-to-date download and installation instructions as well as the Dockerfile can be found at \url{https://github.com/natanielruiz/dockerface}.

At the user's end, after installing NVIDIA drivers, Docker and nvidia-docker, they only have to run a one line command to download and run the Docker container. The very first time the user runs the container, a supplied one line command allows them to compile Caffe, and then the package is ready for all subsequent uses. The user can now process video and images using one line commands which launch the image or video processing scripts included in dockerface.

\section{Reproduction of Accuracy in Widely Used Face Detection Benchmarks.} \label{sec3}
\begin{figure}[t]
\begin{center}
   \includegraphics[width=1\linewidth]{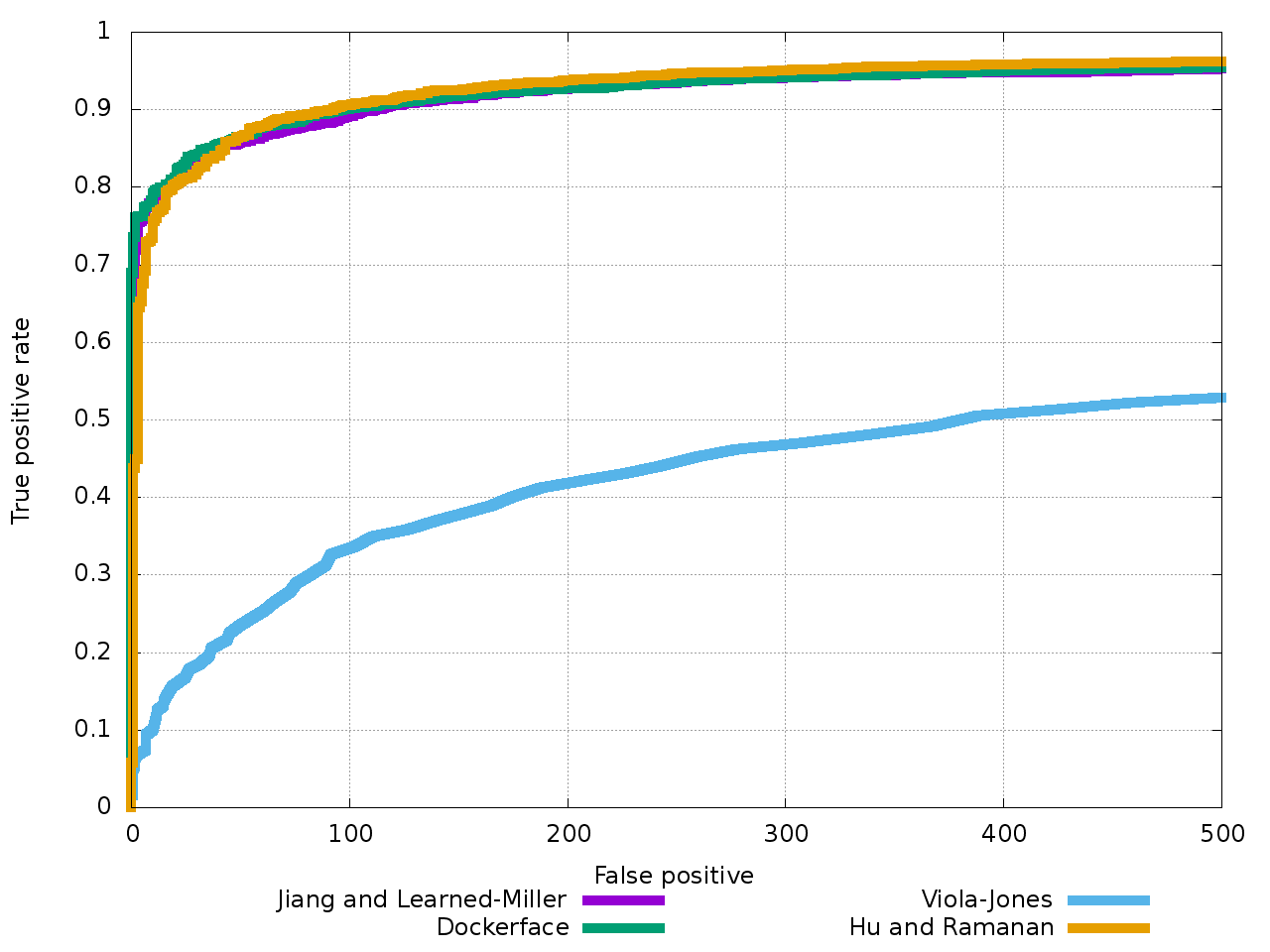}
\end{center}
   \caption{ROC curve on FDDB test dataset comparing Dockerface, Face Faster R-CNN and other widely used standalone detectors.}
   \label{FDDB_ROC}
\end{figure}

We train a Faster R-CNN network with VGG-16~\cite{simonyan2014very} as the detector module on the WIDER Face dataset using for 80,000 iterations using the simultaneous optimization training protocol which trains the Region Proposal Network at the same time as the Faster R-CNN detector module. We use Stochastic Gradient Descent (SGD) with a base learning rate of 0.001, a momentum of 0.9 and weight decay of 0.0005. Note that all of these values are the same as in \cite{jiang2017face}. In essence we replicate experiments first presented in \cite{jiang2017face} after training our own network using their protocol. We obtain almost exact results in the FDDB test dataset, with slight increase in performance which might be due to newer GPU drivers, CUDA and cuDNN software. We present our results in the form of a ROC curve found in Figure \ref{FDDB_ROC} comparing to Jiang and Miller, a very recent state-of-the-art method by Hu and Ramanan~\cite{tinyfaces} and the still widely used Viola-Jones detector~\cite{viola2001rapid} which is implemented in OpenCV.

\section{Does it Work in a Real Research Problem?} \label{sec4}
Here we present a comparison of Dockerface to easy-to-use detectors which are often used in preprocessing steps in research. We annotate around 11,000 frames, or 6 minutes, of video of children in a naturalistic setting interacting with an adult wearing a wearable camera. The annotations consist in bounding boxes around the child faces. This dataset is intended to evaluate the performance of the face detectors in a real research problem. The data comes from children studies conducted at the Georgia Tech Child Study Lab in Atlanta, GA - and has been the center of various research studies such as \cite{ye2015detecting} where Ye et al. obtain the child face bounding box using the OMRON OKAO proprietary detector.

We run several commonplace detectors and Dockerface on these annotated videos and compare the output bounding boxes to the annotated bounding boxes. We show in Table \ref{SIMONS_1} and \ref{SIMONS_2} that the Faster R-CNN network which powers Dockerface is vastly superior to the OpenCV frontal face detector which uses Haar feature-based Cascade Classifiers in the style of Viola-Jones~\cite{viola2001rapid}, Dlib face detector which uses a HoG feature-based linear classifier using spatial pyramids~\cite{dalal2005histograms} and the OMRON OKAO face detector~\cite{omron} - the performance gap exists in both a simpler and more challenging video across different Intersection of Union (IOU) thresholds.

\begin{figure}[t]
\begin{center}
   \includegraphics[width=1\linewidth]{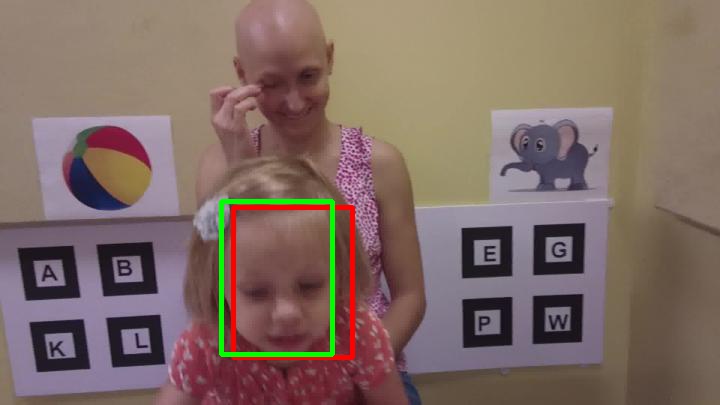}
\end{center}
   \caption{Example frame of a video containing a child participant interacting with a person using a wearable camera. The red bounding box depicts the ground truth annotation while the green bounding box depicts the predicted bounding box by Dockerface. Note that we only display the prediction and annotation of the child's face.}
   \label{SIMONS_FRAME}
\end{figure}

\begin{table}[]
\centering
\caption{Comparison of different face detection methods with Dockerface at different Intersection of Union (IOU) thresholds using around 5,400 frames (3 minutes) from a video containing a child participant interacting with a person using a wearable camera.}
\label{SIMONS_1}
\resizebox{\columnwidth}{!}{%
\begin{tabular}{lrr|rr}
           & \multicolumn{1}{l}{Recall@0.5} & \multicolumn{1}{l|}{Precision@0.5} & \multicolumn{1}{l}{R@0.75} & \multicolumn{1}{l}{P@0.75} \\
OpenCV     & 0.40                           & 0.84                               & 0.03                            & 0.21                               \\
Dlib       & \textbf{0.99}                           & 0.24                               & \textbf{0.98}                            & 0.15                               \\
OMRON      & 0.51                           & 0.95                               & 0.02                            & 0.30                               \\
Dockerface & 0.98                           & \textbf{0.98}                               & 0.93                            & \textbf{0.92}
\end{tabular}%
}
\end{table}

\begin{table}[]
\centering
\caption{Comparison of different face detection methods with Dockerface using around 5,600 frames (3 minutes) from a slightly more challenging child interaction video.}
\label{SIMONS_2}
\resizebox{\columnwidth}{!}{%
\begin{tabular}{lrr|rr}
           & \multicolumn{1}{l}{Recall@0.5} & \multicolumn{1}{l|}{Precision@0.5} & \multicolumn{1}{l}{R@0.75} & \multicolumn{1}{l}{P@0.75} \\
OpenCV     & 0.09                           & 0.83                               & 0.01                            & 0.31                               \\
Dlib       & 0.77                           & 0.36                               & 0.69                            & 0.27                               \\
OMRON      & 0.36                           & 0.93                               & 0.01                            & 0.11                               \\
Dockerface & \textbf{0.91}                  & \textbf{0.95}                      & \textbf{0.73}                   & \textbf{0.82}
\end{tabular}%
}
\end{table}

\section{Related Work} \label{sec5}
The task of object detection, which consists in localizing a bounding box containing an object in an image and classifying this object, has evolved from a sliding window approach combined with hand-crafted features such as SIFT, HOG or Haar features to highly successful convolutional neural network approaches. Faster R-CNN~\cite{ren2015faster} is one of the most widely-used methods for object detection, variants of which have achieved top results in all object detection benchmarks such as Microsoft COCO~\cite{lin2014microsoft} and PASCAL VOC~\cite{everingham2010pascal}. It consists of two modules: a Region Proposal Network (RPN) and a detector module. The RPN is a fully-convolutional neural network which generates object proposals. Using Region of Interest Pooling these proposals are evaluated by the detector module and a classification score is assigned to them which tells us if the region is indeed an object and which object it is.

There has been extensive work in face detection using CNNs which show impressive results in widely used benchmarks such as \cite{jain2010fddb} and \cite{yang2016wider}. A variety of technical avenues have been explored; nevertheless by comparing a Faster R-CNN detector trained on the WIDER Face dataset Jiang and Learned-Miller proved that one can obtain very competitive accuracy using a general purpose object detector without any modifications~\cite{jiang2017face}. They have taken a first step to providing easy face detection for a wider audience by releasing their training and testing code~\cite{face-faster-rcnn} which is based on the great Python Faster R-CNN repository~\cite{python-faster-rcnn}.

Our evaluation of Dockerface leverages a large dataset of children's social interactions with an adult examiner which are available to the research community in the form of the MMDB dataset~\cite{Rehg2013}. We added some additional bounding box annotations of faces to conduct the evaluation, and these annotations are available along with the original videos. See~\url{http://www.cbi.gatech.edu/mmdb/} for details.

\section{Conclusion}
We have replicated work done around face detection using Faster R-CNN and achieved the same accuracy as Jiang and Learned-Miller's Face Faster-RCNN~\cite{jiang2017face} technical report. We have tested several methods in a real task and shown that Faster R-CNN is much more accurate than detectors that are regularly used in research as a pre-processing step such as dlib, OpenCV and OMRON.

Our aim is to replace the use of these detectors in research. We release an easy to download and install Docker container for Linux with a pre-trained Faster R-CNN model for detecting faces accompanied by video and image processing scripts which are easily usable.

{\small
\bibliographystyle{ieee}
\bibliography{egbib}
}

\end{document}